\documentclass[lettersize,journal]{IEEEtran}
\usepackage{amsmath,amsfonts}
\usepackage{algorithmic}
\usepackage{algorithm}
\usepackage{array}
\usepackage[caption=false,font=normalsize,labelfont=sf,textfont=sf]{subfig}
\usepackage{textcomp}
\usepackage{stfloats}
\usepackage{url}
\usepackage{verbatim}
\usepackage{graphicx}
\usepackage{cite}
\usepackage{multirow}
\usepackage{multicol}
\usepackage{booktabs}
\usepackage{bm}
\usepackage{wrapfig}
\usepackage[table,dvipsnames]{xcolor}
\usepackage{pifont}
\usepackage{makecell}

\def\etal{\emph{et al.}}

\newcommand{\cmark}{\ding{51}}%
\newcommand{\xmark}{\ding{55}}%

\usepackage[pagebackref=true,breaklinks=true,letterpaper=true,colorlinks,bookmarks=false]{hyperref}

\begin{document}

\title{ADEM-VL: Adaptive and Embedded Fusion for Efficient Vision-Language Tuning}

\author{Zhiwei~Hao$^*$,
Jianyuan~Guo$^*$,
Li~Shen,
Yong~Luo,
Han~Hu,
        and~Yonggang~Wen
\IEEEcompsocitemizethanks{
  \IEEEcompsocthanksitem Zhiwei Hao and Han Hu are with the School of Information and Electronics, Beijing Institute of Technology, Beijing, China. 
  E-mail: \{haozhw, hhu\}@bit.edu.cn.
  \IEEEcompsocthanksitem Jianyuan Guo is with School of Computer Science, Faculty of Engineering, The University of Sydney, Sydney, Australia. 
  E-mail: jianyuan.guo@sydney.edu.au.
  \IEEEcompsocthanksitem Li Shen is with School of Cyber Science and Technology, Sun Yat-sen University, Shenzhen, China.
  E-mail: mathshenli@gmail.com.
  \IEEEcompsocthanksitem Yong Luo is with School of Computer Science, Wuhan University, Wuhan, China. 
  E-mail: luoyong@whu.edu.cn.
  \IEEEcompsocthanksitem Yonggang Wen is with School of Computer Science and Engineering, Nanyang Technological University, Singapore. 
  E-mail: ygwen@ntu.edu.sg.}
\thanks{Corresponding to Han Hu.}}

\markboth{Adaptive and Embedded Fusion for Efficient Vision-Language Tuning}%
{Hao \MakeLowercase{\textit{et al.}}: Adaptive and Embedded Fusion for Efficient Vision-Language Tuning}


\maketitle

\begin{abstract}
    Recent advancements in multimodal fusion have witnessed the remarkable success of vision-language (VL) models, which excel in various multimodal applications such as image captioning and visual question answering.
    However, building VL models requires substantial hardware resources, where efficiency is restricted by two key factors: the extended input sequence of the language model with vision features demands more computational operations, and a large number of additional learnable parameters increase memory complexity. These challenges significantly restrict the broader applicability of such models.
    To bridge this gap, we propose ADEM-VL, an efficient vision-language method that tunes VL models based on pretrained large language models (LLMs) by adopting a parameter-free cross-attention mechanism for similarity measurements in multimodal fusion. This approach only requires embedding vision features into the language space, significantly reducing the number of trainable parameters and accelerating both training and inference speeds.
    To enhance representation learning in fusion module, we introduce an efficient multiscale feature generation scheme that requires only a single forward pass through the vision encoder.
    Moreover, we propose an adaptive fusion scheme that dynamically discards less relevant visual information for each text token based on its attention score. This ensures that the fusion process prioritizes the most pertinent visual features.
    With experiments on various tasks including visual question answering, image captioning, and instruction-following, we demonstrate that our framework outperforms existing approaches. Specifically, our method surpasses existing methods by an average accuracy of 0.77\% on ScienceQA dataset, with reduced training and inference latency, demonstrating the superiority of our framework. The code is available at \href{https://github.com/Hao840/ADEM-VL}{https://github.com/Hao840/ADEM-VL}.
\end{abstract}


\section{Introduction}
\IEEEPARstart{R}{ecently}, vision-language (VL) modeling has made significant progress~\cite{achiam2023gpt,touvron2023llama}. The main goal of these models is to make predictions based on inputs from both visual and textual data. By leveraging the powerful prompt-following capabilities of pretrained autoregressive large language models (LLMs), fine-tuned VL models achieve remarkable results on various tasks and even surpass well-educated humans in some cases.

Existing approaches for multimodal fusion can be roughly categorized into two categories based on whether the fusion is achieved in the feature space or input space.
The first category fuses visual information into the LLM at intermediate layers~\cite{alayrac2022flamingo}. Specifically, a pretrained CLIP~\cite{radford2021learning} vision tower is used to extract visual features from the input image. 
These features are then aligned with the dimensions of LLMs via a projector. Cross-attention modules facilitate the fusion, with text tokens in the LLM serving as queries and visual features serving as keys and values. Typically, each LLM layer requires a dedicated cross-attention module. These additional modules introduce a significant number of new parameters and increase the overall computational complexity of VL models.
The other category directly fuses vision and text information in the input space of LLMs~\cite{liu2024visual}.
After obtaining visual features extracted by the CLIP model, these methods adopt a learnable projection model to achieve dimension alignment. However, the visual features are directly regarded as input tokens and concatenated at the head of text tokens rather than being fused by complicated schemes. This kind of models require a two-stage training process: in the first stage, only the projector is trained, and in the second stage, both the projector and the LLM are trained together.
Although such models avoid the additional parameters introduced by cross-attention modules, they fine-tune the entire LLM in the second training stage. This results in billions of trainable parameters, demanding substantial storage space and computational resources. Furthermore, the extended input sequence length increases both training and inference costs quadratically.
To enable broader and more cost-effective applications of VL models while minimizing carbon emissions, it is essential for fine-tuned models to be more efficient in terms of both parameters and computation. While previous work~\cite{hu2021lora} on parameter-efficient fine-tuning (PEFT) for pure LLMs can be adapted for VL models, these methods often fail to fully integrate visual information and typically do not achieve optimal performance.

To bridge this gap, we propose ADEM-VL, an efficient \textbf{ad}aptive and \textbf{em}bedded fusion framework for \textbf{v}ision-\textbf{l}anguage tuning at the intermediate layers of pretrained LLMs.
To reduce the number of parameters in the cumbersome cross-attention modules of existing VL models, we first explore their simplified variants. By reformulating standard cross-attention into an abstract form, we replace the parameterized similarity measurement with a parameter-free approach. This eliminates most of the learnable parameters in these modules, except for a shared low-rank projector that aligns the dimensions for embedding vision features.
We further enhance the representation learning ability of our parameter-free fusion module by introducing multiscale visual features through pooling and concatenation operations. Unlike existing approaches that require multiple invocations of the vision encoder~\cite{liu2024improved,xu2024llava}, our method only requires a single forward pass, resulting in negligible additional computational cost.
Furthermore, considering that not all image features contribute equally to prediction, we propose an adaptive fusion scheme. For each text token, the corresponding image features with lower attention scores are dropped. This allows the text tokens to focus on more relevant visual features and avoid interference from irrelevant information.
These designs significantly reduce the learnable parameters in both tuning and inference stage while requiring no increase in input length. With enhanced multimodal fusion schemes, our ADEM-VL achieves superior performance while maintaining high efficiency.
Our method differs from existing methods that introduce adapter modules into LLMs and fuse multimodal information in the input space~\cite{zhang2023llama,luo2024cheap}. While these methods consider parameter efficiency, they often ignore computational complexity. In contrast, our framework can achieve both parameter and computationally efficient fusion for VL models. 

To evaluate the proposed framework, we conduct experiments on three different vision-language tasks.
On the visual question answering task using ScienceQA dataset, our fine-tuned LLaMA-13B model achieves 94.55\% average accuracy, outperforming existing approaches by 0.77\% while being 15\% and 3\% faster in training and inference stage, respectively.
Additionally, the results on the image captioning task and the instruction-following task demonstrate comparable performance to existing methods, further validating the effectiveness of our proposed framework.
The contributions of our paper can be summarized as follows:
\begin{itemize}
    \item We propose ADEM-VL, an adaptive and embedded fusion framework for vision-language tuning. ADEM-VL is highly efficient in terms of both parameters and computational cost during training and inference.
    \item By reformulating the standard cross-attention and replacing the parameterized similarity measurement with a parameter-free one, we significantly reduce the number of trainable parameters.
    \item We introduce pooling and concatenation operations to generate multiscale visual features with a single invocation of vision encoder, resulting in negligible additional computational cost.
    \item We implement an adaptive fusion scheme that discards irrelevant image features with lower attention scores, enabling text tokens to focus on more pertinent visual information and reducing interference.
\end{itemize}

\section{Related work}

\subsection{Multimodal architectures.}

The advancements in the field of multimodal learning have been significantly shaped by breakthroughs in natural language processing, particularly with the advent of attention-based models~\cite{vaswani2017attention}. 
Inspired by BERT, numerous studies have incorporated masked modeling to develop multimodal systems~\cite{chen2020uniter,gan2020large,li2020hero,li2020oscar,lu2019vilbert,singh2022flava,zhu2020actbert,su2019vl,tan2019lxmert,zhang2024open}. Additionally, some multimodal models have employed contrastive learning techniques during their training processes~\cite{bain2021frozen,jia2021scaling,li2021align,miech2020end,radford2021learning,yuan2021florence}. Both approaches have demonstrated remarkable performance across a range of multimodal tasks.

Recently, driven by the success of LLMs in language, researchers have increasingly concentrated on developing autoregressive multimodal models to harness the exceptional capabilities of LLMs.
A prior work, Flamingo~\cite{alayrac2022flamingo}, bridged the gap between pure language models and vision models through a cross-attention mechanism. Specifically, this model integrates gated cross-attention layers before each block of the language model, facilitating interaction between vision and language modalities. 
Flamingo surpassed previous models and showcased the potential of autoregressive architectures for multimodal learning. This achievement inspired a series of subsequent VL models, which can be broadly categorized into two classes based on how they incorporate vision information into the pretrained language model.

One class of models integrates vision information into the internal layers of the language model.
Within this category, vision information is first extracted using a pretrained vision encoder, followed by a learnable projection layer or a resampling layer.
The multimodal fusion is then achieved by inserting cross-attention layers into the LLMs, either before the the self-attention layers~\cite{alayrac2022flamingo,li2023mimic} or after the self-attention layers~\cite{cho2021unifying}.
In addition to standard cross-attention layers, some architectures employ customized layers specifically designed for fusion~\cite{zhang2023llama,gao2023llama,wang2023cogvlm,jia2024geminifusion,dong2024internlm}.

Another class of models introduces vision information at the input stage of the language model.
In this approach, the extracted vision information is directly concatenated with the language tokens before being input into the LLMs.
Typically, a projection layer is employed to align the vision feature space with the language token space.
This projection layer can be a simple linear layer~\cite{liu2024visual,liu2024improved,driess2023palm,chen2023minigpt}, a resampling layer~\cite{li2023blip,zhu2023minigpt}, or other customized layers designed to facilitate effective fusion~\cite{mckinzie2024mm1,ye2023mplug,mu2024embodiedgpt,maaz2023video,bai2023qwen}.
Additionally, some works use tokenized source images as input instead of extracted features~\cite{jin2023unified,lu2022unified}.
This design enables their capability for image generation.

To facilitate a better understanding of VL model performance, various benchmarks and toolkits have been introduced, such as MME~\cite{fu2023mme}, MMB~\cite{liu2023mmbench}, MMMU~\cite{yue2024mmmu}, MMT~\cite{ying2024mmt}, and AVIBench~\cite{zhang2024avibench}.  
These benchmarks have driven significant advancements in the architectures of VL models.

Although existing multimodal architectures have achieved remarkable performance, inserting cross-attention modules at intermediate layers of LLMs introduces a substantial number of additional trainable parameters.
On the other hand, concatenating vision features with language tokens extends the length of the input sequence, significantly increasing the computational resources needed for inference.
Consequently, developing more efficient multimodal architectures is appealing.

\subsection{Parameter-efficient fine-tuning}

Since the advent of LLMs, there has been a sharp increase in the number of model parameters.
While these models deliver outstanding performance, fine-tuning them for downstream tasks is challenging due to their massive size.
To address this issue, various PEFT approaches have been developed.

PEFT techniques for language models can be broadly classified into two categories based on whether they introduce new trainable parameters into the pretrained models.
The first category includes approaches that do not introduce any new parameters.
The main idea behind these methods is to selectively fine-tune the existing parameters within the original model~\cite{zaken2021bitfit,guo2020parameter,sung2021training}.
For example, BitFit~\cite{zaken2021bitfit} focuses on training only the bias values in the pretrained model, significantly reducing the number of trainable parameters required during fine-tuning.
The second category of PEFT techniques involves adding additional trainable parameters while keeping the pretrained model fixed~\cite{houlsby2019parameter,hu2021lora,he2021towards,karimi2021compacter,liu2022few,sung2022lst}.
Adapter~\cite{houlsby2019parameter} and LoRA~\cite{hu2021lora} are two typical methods in this category.
The Adapter approach introduces small linear layers after each sub-block of the pretrained model and makes only these new layers trainable. LoRA uses low-rank matrix decomposition to parameterize the pretrained weight matrices. During inference, these decomposed matrices can be merged back into the original model, thus incurring no additional inference cost.

Generally, most PEFT approaches designed for language models can be directly applied to the training of multimodal architectures. In addition to these, there are also PEFT methods specifically tailored for multimodal learning~\cite{sung2022vl,hu2023vl,zhang2023llama,gao2023llama,luo2024cheap,jie2024memory}.
VL-Adapter~\cite{sung2022vl} and VL-PET~\cite{hu2023vl} introduce vision information at the input stage and fine-tune pretrained language models with an encoder-decoder architecture using adapter modules. 
LLaMA-Adapter~\cite{zhang2023llama,gao2023llama} and LaVIN~\cite{luo2024cheap} focus on PEFT for multimodal learning using the more advanced LLaMA architecture by designing specific adapter modules.
MenVP~\cite{jie2024memory} accomplishes multimodal fusion by augmenting the linear layers of the pretrained language model with vision information. 

However, while these approaches effectively reduce the number of learnable parameters in multimodal models, many of them extend the input sequence length by combining vision information with language tokens, leaving the challenge of high computational complexity during inference unresolved.
Multimodal fusion at intermediate layers via cross-attention incurs lower computational complexity compared to extending input sequences. Yet, parameter-efficient techniques for optimizing cross-attention modules remain largely unexplored. 
In this paper, we aim to bridge this gap by developing methods that achieve both parameter and computational efficiency in multimodal fusion.

\begin{figure*}
    \centering
    \includegraphics[width=\textwidth]{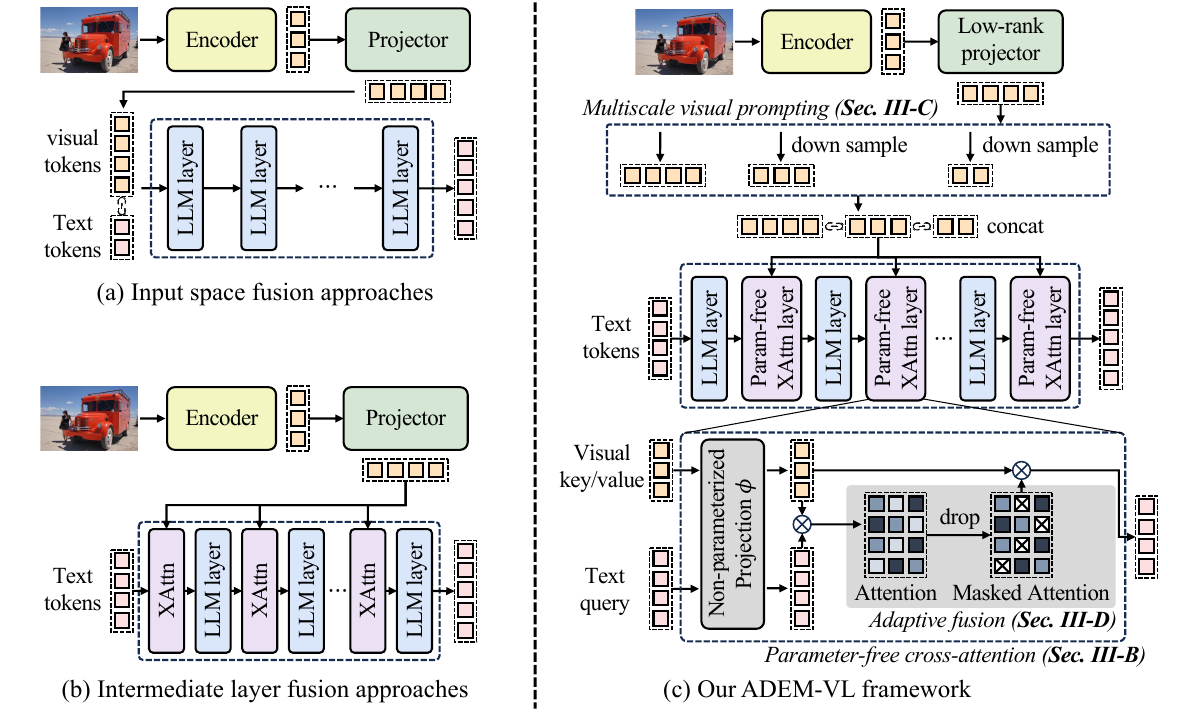}
    \caption{Comparison of different vision-language tuning frameworks: (a) Methods that directly extend the input space of the language model with extracted vision features. (b) Methods that fuse vision information into the language model via cross-attention. (c) Our proposed ADEM-VL framework, which incorporates parameter-free cross-attention, multiscale visual prompting, and adaptive multimodal fusion designs. This approach ensures both parameter and computational efficiency while delivering promising performance.}
    \label{fig:main}
\end{figure*}

\section{Method}

To achieve effective and efficient multimodal fusion for VL models, we propose a framework named ADEM-VL, as shown in Figure~\ref{fig:main}.
To ensure efficiency in both tuning and inference stage, we simplify the standard cross-attention module through removing trainable parameters and thus reduce its computational requirements.
To deliver promising performance with the simplified fusion scheme, we further introduce an effective multiscale visual information generating scheme to provide the language model with ample visual knowledge and an adaptive fusion scheme to help the model focus only on informative visual information.
In this section, we first introduce the related background knowledge (\ref{sec:method:background}), followed by the simplified parameter-free cross-attention (\ref{sec:method:linearize}), the multiscale prompting (\ref{sec:method:multiscale}), and the adaptive fusion (\ref{sec:method:adaptive}). Finally, we present the overall framework of our ADEM-VL based on these proposed designs (\ref{sec:method:framework}).

\subsection{Background}\label{sec:method:background}

\textbf{Vision-Language models.}
VL models are designed to process both language and visual inputs to perform multimodal inference tasks such as visual question answering (VQA). Recently, due to the tremendous success of LLMs, there have been efforts to leverage these exceptional models to build VL systems.
Generally, there are two main classes of approaches for integrating vision information into pretrained LLMs.
One class of methods achieves multimodal fusion by adopting cross-attention to merge LLM features and vision features at intermediate layers. The other class of methods concatenates vision tokens with text tokens at the input space. However, these approaches either introduce considerable additional parameters or increase inference costs due to the introduction of new modules or the extension of input length.
Compared to directly increasing the input length, multimodal fusion via cross-attention offers greater potential for optimizing efficiency, as computational complexity increases quadratically with input length. Therefore, we adopt cross-attention-based fusion approaches as the starting point to explore a more efficient multimodal fusion method for VL models.

\textbf{Cross-attention for fusion.}
Existing methods that use cross-attention for visual-language fusion typically employ the language token $\bm{X}_l\in \mathbb{R}^{L\times d}$ as the query and the visual features $\bm{X}_v \in \mathbb{R}^{N\times d'}$ as the key and value to facilitate information interaction~\cite{alayrac2022flamingo,li2022blip}.
This process can be formulated as:
\begin{equation}
    \begin{aligned}
        \text{XAttn}&(\bm{X}_l, \bm{X}_v) = \text{softmax}\left(\frac{\bm{Q} \bm{K}^T}{\sqrt{d_k}}\right) \bm{V}\\
        &=\text{softmax}\left(\frac{\bm{X}_l \bm{W}_Q  \bm{W}_K^T \bm{X}_v^T}{\sqrt{d_k}}\right) \bm{X}_v \bm{W}_V \bm{W}_o^T,
    \end{aligned}
    \label{eq:xattn}
\end{equation}
where $\bm{W}_Q$, $\bm{W}_K$, and $\bm{W}_V$ are learned projection matrices for the query, key, and value, respectively. 
$\bm{W}_o^T$ is the output projection matrix, and $d_k$ is the dimensionality of each key vector.

Following Equation~\ref{eq:xattn}, the adoption of cross-attention presents a remarkable drawback, where each cross-attention module contains four projection matrices, introducing a substantial number of additional trainable parameters. This is particularly significant when multimodal fusion is performed at each layer of the pretrained LLM using individual cross-attention modules~\cite{alayrac2022flamingo}. 
To improve the efficiency of the fusion process, we need to take a closer look at the cross-attention mechanism and modify it to be more efficient.

\subsection{Parameter-free cross-attention}\label{sec:method:linearize}

To improve the efficiency of cross-attention in VL models, we begin our analysis by obtaining an abstract form of Equation~\ref{eq:xattn}.
In the cross-attention module, each query vector performs a dot product with all key vectors, followed by a softmax function to obtain the attention scores. The value vectors are then weighted and summed using these scores and subsequently multiplied by the output projection matrix to obtain the fused feature.
Thus, Equation~\ref{eq:xattn} can also be formulated as:
\begin{equation}
    \text{XAttn}(\bm{X}_l, \bm{X}_v)_i = \frac{\sum_{j}\text{sim}(\bm{Q}_i,\bm{K}_j)\bm{V}_j}{\sum_{j}\text{sim}(\bm{Q}_i,\bm{K}_j)},
\end{equation}
where $\text{sim}(\bm{q},\bm{k})=\exp({\frac{\bm{q}^T\bm{k}}{\sqrt{d_k}}})$ computs similariry between query vector $\bm{q}$ and key vector $\bm{k}$.
Generally, we can define different $\text{sim}(\bm{q}, \bm{k})$ functions to obtain variants of standard cross-attention, such as polynomial attention, RBF kernel attention~\cite{tsai2019transformer}, or linear attention~\cite{katharopoulos2020transformers}. 
Since $\text{sim}(\cdot, \cdot)$ acts as a similarity metric, the only constraint is that it must produce non-negative values. Therefore, it can be any kernel function $k(x, y) : \mathbb{R}^{\|x\| + \|y\|} \to \mathbb{R}_+$.

Based on kernel trick, the kernel function can be defined as $k(x, y)=\phi(x)\phi(y)^T$, where $\phi$ is a projection function.
On the other hand, in the general form of cross-attention, we have: $\text{sim}(\bm{Q}_i,\bm{K}_j)=\exp({\frac{\bm{X}_l \bm{W}_Q  \bm{W}_K^T \bm{X}_v^T}{\sqrt{d_k}}})$.
By selecting an appropriate non-parameterized projection function $\phi$, we can eliminate the need for parameters in the matrices $\bm{W}_Q$ and $\bm{W}_K$.
A possible class of such projections includes the modern activation function ReLU~\cite{hahnloser2000digital} and its modifications like GeLU~\cite{hendrycks2016gaussian} and SiLU~\cite{elfwing2018sigmoid}. These are pointwise projections that can be computed efficiently while preserving strong representation extraction capabilities.
To achieve better performance, variants of ReLU may output negative values when the input is close to zero. However, we find that this limitation has little effect on performance, and the SiLU activation function can serve as an adequate replacement for $\phi$.

Then, to further improve the parameter efficiency of the cross-attention module, we adopt identity matrices for the value vector projection and the output projection.
This is based on the assumption that the powerful pretrained LLMs can work effectively with coarsely fused information to accomplish multimodal tasks.
Finally, the embedded parameter-free cross-attention can be formulated as:
\begin{equation}
    \text{XAttn}(\bm{X}_l, \bm{X}_v)=\phi(\bm{X}_l) \phi(\bm{X}_v)^T \bm{X}_v,
    \label{eq:xattn_ours}
\end{equation}
where $\phi(\cdot)=\text{SiLU}(\cdot)$.

In VL models, the original feature sizes of the vision tower and the LLM are usually not the same. 
Existing work~\cite{liu2024visual} adopts a learnable projection matrix to align their dimensions.
We follow a similar idea but decompose the projection matrix into low-rank decompositions to reduce the parameter count.
This process embeds visual features into the language feature space, enabling parameter-free attention-based fusion, which explains the ``embedded'' aspect in our method's name.
Note that the dimension alignment serves as a pre-processing step for visual features, meaning that the aligned features are directly utilized in parameter-free cross-attention modules across all layers of the pretrained LLM. As a result, we do not count its parameters as part of the cross-attention modules.

\textbf{computational complexity.}
Suppose the dimension of feature vectors is $d$. In a standard cross-attention module with $L$ query and $N$ key/value vectors, the FLOPs of each original cross-attention module is $2Ld^2+2Nd^2+2LNd$, while in our simplified module, the FLOPs become $2LNd$.
Since LLMs usually have very large hidden dimensions\footnote{For example, both LLaMA-7B and LLaMA2-7B have a hidden dimension of $d=4096$ and support input lengths of $L=2048$ and $L=4096$, respectively. However, in most VL settings, the input length $l$ is usually much smaller than $L$ ($l<<L$).} $d$, the removal of the matrix projection requirement greatly reduces the computational complexity by approximately $2Ld^2+2Nd^2$ FLOPs.
Hence, our simplification of the cross-attention module helps improve both parameter and computational efficiency, facilitating PEFT of VL models.

\subsection{Multiscale visual prompts}\label{sec:method:multiscale}

Multiscale features have been proven effective for improving model performance in various computer vision tasks~\cite{lin2017feature,hao2022cdfkd}.
Given its success, we believe that introducing multiscale vision information in VL models could also help improve model performance.

To obtain multiscale visual information, we generate features of different scales by pooling operation based on the extracted features of the original image.
Take CLIP encoder with an input resolution of 224$\times$224 as an example, the input image is split into 256 patches, resulting in 256 extracted visual features, each corresponding to an original image patch. To provide the language model with high-level visual features, we merge adjacent tokens by pooling.
Specifically, we first reshape the 256 one-dimensional tokens into a 16$\times$16 two-dimensional grid based on their positions in the original image. Then, we apply pooling operations with different kernel sizes to obtain multiscale visual features. By flattening the feature grid of each scale and concatenating them together, we obtain the final visual feature for LLMs.
We compare different pooling configurations in terms of kernel type and size in our experiments.

In addition to the extracted features, the CLIP models also output a [cls] token, which acts as a more abstract global representation of the input image. To further utilize this token, we follow the input space prompting approaches~\cite{liu2024visual} by concatenating it at the head of the text tokens. Unlike existing works that extend input length by dozens of tokens, we introduce only one additional token, which incurs negligible extra inference overhead, despite the computational complexity increasing quadratically with input length.

\textbf{Discussion.}
The main restriction on the application of multiscale vision features on multimodal modeling is the limited availability of pretrained visual encoder models. Currently, CLIP is the most frequently used encoder.
However, the publicly available well-pretrained CLIP models only support input sizes of 224$\times$224 or 336$\times$336, preventing users from using inputs of other sizes.
Existing works have explored designing approaches to inject multiscale vision information into language models~\cite{liu2024improved,xu2024llava}.
To bridge this gap, these works mainly divide and resize the original image and then process each patch individually to obtain multiscale visual features.
For instance, Liu \etal~\cite{liu2024improved} divided the original image into 2$\times$2 sub-images and extract features from each resized sub-image. Then, features of all sub-images and the original image are concatenated along with text tokens as the input of LLMs.
Xu \etal~\cite{liu2024improved} designed a more flexible method that can split image dynamically.
However, these approaches require multiple forward passes through the vision encoder to process all sub-images individually, without considering their relationships within the entire image.
Furthermore, while these methods focus on achieving multimodal fusion in the input space, the use of multiscale information for cross-attention-based fusion remains underexplored.
In comparison, our design generates features of different scales without invoking the vision encoder multiple times, thus maintaining efficiency.

\subsection{Adaptive multimodal fusion}\label{sec:method:adaptive}

The introduction of multiscale visual features provides the language model with more detailed information about the input image. However, not all of this information may be useful for each text query. For instance, for an inquiry about the left part of the input image, features corresponding to the right part of the image would be useless or could even mislead the model, preventing it from making correct predictions.

To address this challenge, we design an adaptive multimodal fusion scheme to help the text tokens dynamically extract visual information in the parameter-free cross-attention module. This allows them to focus more on informative visual features and mitigate interference from irrelevant ones.
Following Equation~\ref{eq:xattn_ours}, the attention score in our modified cross-attention is $\phi(\bm{X}_l) \phi(\bm{X}_v\bm{W}')^T \in \mathbb{R}^{L\times N}$. 
This matrix indicates the similarity score between each text token and each visual feature.
Here we assume that for each text token, the visual features corresponding to larger similarity scores are more important for prediction, while those with lower scores provide less or even useless information. Based on this assumption, we can drop the less informative visual features to help improve performance.
Specifically, we sort the attention score matrix over each row individually and mask the lowest values with a mask ratio of $\gamma$, indicating that the visual features with low scores will be dropped.
As there is no restriction for the summation of attention scores for each text token to be 1, unlike the softmax function, dropping some visual features does not significantly change the attention mechanism in our modified cross-attention.
By generating the mask individually for each row of the score matrix, the drop decision is adaptive to different text tokens and input images.
This allows each text token to focus more on useful information, helping to achieve better fusion results.

\subsection{Overall framework}\label{sec:method:framework}

\includegraphics[width=0.9\linewidth]{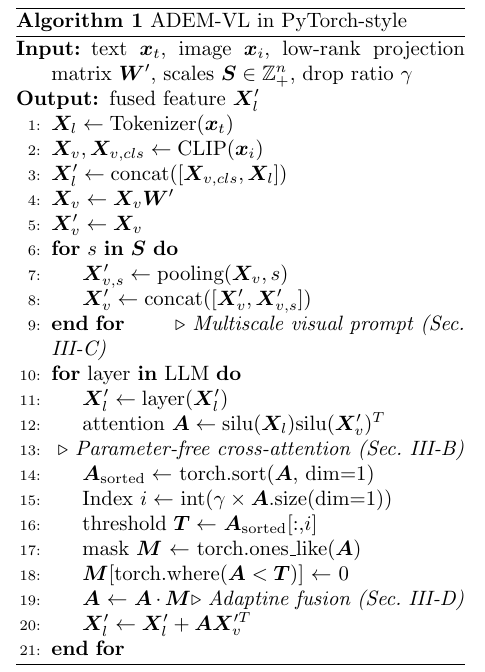}

With the parameter-free cross-attention, multiscale visual features, and adaptive fusion scheme, we can build our adaptive and embedded fusion framework (ADEM-VL).
Specifically, following the common cross-attention-based framework~\cite{alayrac2022flamingo}, we insert our modified non-parameterized cross-attention module at each layer of the target LLM.
To align the dimension of visual feature and the [cls] token with that of the LLM while ensuring parameter efficiency, we adopt two low-rank projection matrices.
Moreover, we adopt the idea of introducing lightweight adapters into the vision tower of CLIP and adding learnable positional embeddings $\bm{E}$ to the visual features~\cite{jie2024memory}, where the embeddings are shared across all cross-attention modules.
The projection matrices, adapters for CLIP, and learnable positional embeddings consist of all trainable parameters in our framework.
Finally, the fused feature can be denoted as:
\begin{equation}
    \bm{X}'=\alpha \text{XAttn}(\bm{X}_l, \beta\bm{X}_v + \bm{E})
    \label{eq:overall}
\end{equation}
where $\alpha$ and $\beta$ are weighting hyperparameters for visual features and fused features.
We compare the impact of these parameters in our experiments.

We provide details of our proposed ADEM-VL framework in PyTorch-style in Algorithm 1.
Note that we add cross-attention modules after each LLM layer in this algorithm for clarification.
In practice, we can insert these modules in the middle of LLM layers, and different configurations are compared in our experiments.

\begin{table*}
    \renewcommand\tabcolsep{3pt}
    \centering
    \small
    \caption{Evaluation results on ScienceQA test set. NAT = natural science, SOC = social science, LAN = language science, TXT = text context, IMG = image context, NO = no context, G1-6 = grades 1-6, G7-12 = grades 7-12.}
    \begin{tabular}{lcc|ccc|ccc|cc|c}
        \toprule
        \multirow{2}{*}{Method}                          & \multicolumn{2}{c|}{\#Param} & \multicolumn{3}{c|}{Subject} & \multicolumn{3}{c|}{Context Modality} & \multicolumn{2}{c|}{Grade} & \multirow{2}{*}{Average}                                                                                                       \\
                                                         & Trainable & LLM & NAT            & SOC            & LAN            & TXT            & IMG            & NO             & G1-6           & G7-12          &                \\
        \midrule
        \multicolumn{2}{l}{\emph{Zero-/few-shot methods}}                                                                                                                                                                           \\
        Human~\cite{lu2022learn}                         & -         & -   & 90.23          & 84.97          & 87.48          & 89.60          & 87.50          & 88.10          & 91.59          & 82.42          & 88.40          \\
        GPT-3.5~\cite{lu2022learn}                       & -         & -   & 74.64          & 69.74          & 76.00          & 74.44          & 67.28          & 77.42          & 76.80          & 68.89          & 73.97          \\
        GPT-3.5~\cite{lu2022learn}                       & -         & -   & 75.44          & 70.87          & 78.09          & 74.68          & 67.43          & 79.93          & 78.23          & 69.68          & 75.17          \\
        GPT-4~\cite{achiam2023gpt}                       & -         & -   & 84.06          & 73.45          & 87.36          & 81.87          & 70.75          & 90.73          & 84.69          & 79.10          & 82.69          \\
        \midrule
        \multicolumn{2}{l}{\emph{Full training methods}}                                                                                                                                                                            \\
        UnifiedQA~\cite{lu2022learn}                     & 223M      & -   & 71.00          & 76.04          & 78.91          & 66.42          & 66.53          & 81.81          & 77.06          & 68.82          & 74.11          \\
        MM-CoT$_\text{Base}$~\cite{zhang2023multimodal}  & 223M      & -   & 87.52          & 77.17          & 85.82          & 87.88          & 82.90          & 86.83          & 84.65          & 85.37          & 84.91          \\
        MM-CoT$_\text{Large}$~\cite{zhang2023multimodal} & 733M      & -   & 95.91          & 82.00          & 90.82          & 95.26          & 88.80          & 92.89          & 92.44          & 90.31          & 91.68          \\
        LLaVA~\cite{liu2024visual}                       & 7B        & 7B  & -              & -              & -              & -              & -              & -              & -              & -              & 89.84          \\
        LLaVA~\cite{liu2024visual}                       & 13B       & 13B & 90.36          & 95.95          & 88.00          & 89.49          & 88.00          & 90.66          & 90.93          & 90.90          & 90.92          \\
        \midrule
        \multicolumn{2}{l}{\emph{PEFT methods with LLaMA}}                                                                                                                                                                          \\
        LLaMA-Adapter~\cite{zhang2023llama}              & 1.8M      & 7B  & 84.37          & 88.30          & 84.36          & 83.72          & 80.32          & 86.90          & 85.83          & 84.05          & 85.19          \\
        LLaVA-LoRA~\cite{jie2024memory}                  & 4.4M      & 7B  & 91.70          & 94.60          & 86.09          & 91.25          & 90.28          & 88.64          & 91.52          & 89.65          & 90.85          \\
        LaVIN~\cite{luo2024cheap}                        & 3.8M      & 7B  & 89.25          & 94.94          & 85.24          & 88.51          & 87.46          & 88.08          & 90.16          & 88.07          & 89.41          \\
        LaVIN~\cite{luo2024cheap}                        & 5.4M      & 13B & 90.32          & 94.38          & 87.73          & 89.44          & 87.65          & 90.31          & 91.19          & 89.26          & 90.50          \\
        MemVP~\cite{jie2024memory}                       & 3.9M      & 7B  & 94.45          & 95.05          & 88.64          & 93.99          & 92.36          & 90.94          & 93.10          & 93.01          & 93.07          \\
        MemVP~\cite{jie2024memory}                       & 5.5M      & 13B & 95.07          & 95.15          & 90.00          & 94.43          & 92.86          & 92.47          & 93.61          & 94.07          & 93.78          \\
        \rowcolor[gray]{0.9}
        ADEM-VL                                          & 4.5M      & 7B  & 95.52          & \textbf{95.39} & 89.18          & 95.36          & \textbf{93.95} & 90.94          & 93.87          & 93.80          & 93.85          \\
        \rowcolor[gray]{0.9}
        ADEM-VL                                          & 5.5M      & 13B & \textbf{96.00} & 94.94          & \textbf{91.27} & \textbf{95.45} & \textbf{93.95} & \textbf{93.03} & \textbf{94.46} & \textbf{94.73} & \textbf{94.55} \\
        \midrule
        \multicolumn{2}{l}{\emph{PEFT methods with LLaMA2}}                                                                                                                                                                         \\
        MemVP~\cite{jie2024memory}                       & 3.9M      & 7B  & 93.12          & 94.60          & 89.27          & 92.86          & 91.13          & 91.15          & 92.51          & 92.29          & 92.43          \\
        \rowcolor[gray]{0.9}
        ADEM-VL                                          & 4.5M      & 7B  & \textbf{95.74} & \textbf{94.83} & \textbf{90.00} & \textbf{95.50} & \textbf{93.75} & \textbf{91.78} & \textbf{94.16} & \textbf{93.87} & \textbf{94.06} \\
        \bottomrule
    \end{tabular}
    \label{tab:main}
\end{table*}

\section{Experiment}

To evaluate the effectiveness of our ADEM-VL framework, we conduct a series of experiments on various datasets.
In this section, we first introduce our experimental setup (\ref{sec:exp:setup}).
Then, we provide a quantitative comparison between different VL model tuning approaches (\ref{sec:exp:quan}), followed by an ablation study to validate each component in our framework (\ref{sec:exp:ablation}).
Finally, we present qualitative results of our trained model (\ref{sec:exp:qual}).

\subsection{Experimental setup}\label{sec:exp:setup}

\textbf{Datasets.}
We conduct experiments primarily on the ScienceQA dataset~\cite{lu2022learn}, a multimodal dataset for science question answering. The dataset consists of text-only and image-text samples covering 3 subjects, 26 topics, and 379 skills. These samples are split into train, validation, and test sets. We report the average accuracy on the test set.
We also evaluate our method on the COCO Caption dataset~\cite{chen2015microsoft} using the Karpathy split of the dataset~\cite{karpathy2015deep}.
We further train the model with instruction-following datasets. Specifically, we adopt Alpaca-52K~\cite{alpaca}, a text-only dataset generated by GPT-3.5~\cite{brown2020language}, and LLaVA-158K~\cite{liu2024visual}, a text-image dataset generated by GPT-4~\cite{achiam2023gpt}. 
To evaluate the zero-shot performance of the trained model, we utilize several image question answering datasets and image understanding benchmarks, including VQAv2~\cite{goyal2017making}, GQA~\cite{hudson2019gqa}, MME~\cite{fu2023mme}, MMB~\cite{liu2023mmbench}, and MMMU~\cite{yue2024mmmu}.

\textbf{Models.}
Our framework fine-tunes pretrained language models for VL tasks.
To achieve this, we adopt LLaMA~\cite{touvron2023llama} as the language model, which is a series of foundation language models ranging from 7B to 65B parameters. Here, we mainly conduct experiments using the variants containing 7B and 13B parameters.
To obtain visual information, we adopt the CLIP~\cite{radford2021learning} model with the ViT-L/14~\cite{dosovitskiy2020image} backbone as the image encoder.

\textbf{Optimization.}
The optimization configuration of our model on each dataset follows that of LaVIN~\cite{luo2024cheap}.
Specifically, we train the model for 20, 5, and 15 epochs on the ScienceQA, COCO Caption, and instruction-following datasets, respectively, with a global batch size of 32.
The learning rate is initialized at 9e-3 and decreases according to a cosine schedule.
We adopt features of two scales as the multiscale visual prompt: the original features and 2$\times$ downsampled features using average pooling. Across all our experiments, hyperparameters $\alpha$, $\beta$, and $\gamma$ are set to 0.1, 0.01, and 0.2, respectively.

\subsection{Quantitative results}\label{sec:exp:quan}

\textbf{ScienceQA.}
We first evaluate our proposed ADEM-VL framework on the ScienceQA dataset and report the results in Table~\ref{tab:main}.
The baseline results are obtained from~\cite{luo2024cheap, jie2024memory}.
From the results, our method achieves the best average accuracy among all the compared methods.
Overall, PEFT methods outperform zero- or few-shot methods and regular methods that fine-tune the whole LLM.
Within PEFT approaches, our ADEM-VL achieves better performance than the best performed baselines while maintaining a similar number of trainable parameters.
Specifically, with LLaMA-7B as the LLM, our method achieves 93.85\% average accuracy, outperforming the second-best baseline by 0.78\%, with only 0.6M more parameters.
When the LLM is LLaMA-13B, the average accuracy of our method increases to 94.55\%, which is 0.77\% higher than the second-best result.
Note that unlike the implementation in MemVP~\cite{jie2024memory}, where LLaMA-13B uses longer positional embeddings than LLaMA-7B and introduces more additional parameters, the trainable parameters in our method increase more slowly with the increase of LLM size compared to MemVP.

Additionally, we conduct experiments using the more advanced pretrained LLM, LLaMA2-7B~\cite{touvron2023llama}, applying the same VL tuning recipe as with LLaMA-7B.  
MemVP, when based on LLaMA2-7B, exhibits poorer performance compared to its LLaMA-7B counterpart.  
In contrast, our ADEM-VL method outperforms the LLaMA-7B variant, highlighting its potential to achieve even superior performance when equipped with more powerful pretrained LLMs.
To ensure a fair comparison with existing approaches, which predominantly use LLaMA models as the pretrained LLM, we adhere to this setting in all subsequent experiments.

\begin{table}
    \renewcommand\tabcolsep{3pt}
    \centering
    \small
    \caption{Evaluation results on COCO caption using the Karpathy test split with LLaMA-13B as the language model. \#T. = trainable parameters. *PEFT methods.}
    \begin{tabular}{lccc}
        \toprule
        Method                                   & \#T. & BLEU-4 & CIDEr \\
        \midrule
        ClipCap~\cite{mokady2021clipcap}         & -    & 33.5   & 113.1 \\
        VisionLLM-H~\cite{wang2024visionllm}     & -    & 32.1   & 114.2 \\
        BLIP~\cite{li2022blip}                   & 583M & 40.4   & 136.7 \\
        BLIP-2~\cite{li2023blip}                 & 188M & 43.7   & 145.3 \\
        $^*$LLaMA-Adapter V2~\cite{gao2023llama} & 14M  & 36.2   & 122.2 \\
        $^*$LaVIN~\cite{luo2024cheap}            & 5.4M & 37.8   & 131.7 \\
        \rowcolor[gray]{0.9}
        $^*$ADEM-VL                              & 5.5M & 38.5   & 133.2 \\
        \bottomrule
    \end{tabular}
    \label{tab:caption}
\end{table}

\textbf{COCO caption.}
We further evaluate the proposed framework on the image captioning task using the COCO caption dataset, following the implementation of BLIP~\cite{li2022blip} and baseline results in LaVIN~\cite{luo2024cheap}.
As shown in the results in Table~\ref{tab:caption}, the proposed method achieves comparable performance while requiring significantly fewer trainable parameters compared to large-scale pre-training approaches such as BLIP~\cite{li2022blip} and BLIP-2~\cite{li2023blip}. Among PEFT approaches, our ADEM-VL outperforms existing baselines by a significant margin, with a 0.7 improvement in the BLEU-4 score and a 1.5 increase in the CIDEr score. These results further demonstrate the superiority of the proposed VL fusion framework.

\begin{table*}
    \renewcommand\tabcolsep{5pt}
    \centering
    \small
    \caption{Evaluation results on the MME benchmark with LLaMA-13B as the language model. MME-C and MME-P measure the perception and cognition abilities of the model, respectively. Extra tokens refer to the number of additional tokens processed by the LLM beyond the standard text tokens. \#T. = trainable parameters. *PEFT methods.}
    \begin{tabular}{lcccc}
        \toprule
        Method                                              & \#Trainable param & \#Extra tokens & MME-P           & MME-C           \\
        \midrule
        LLaVA~\cite{liu2024visual}                          & 13B               & 256            & 502.8           & 214.6           \\
        $^*$Prompt-Aware Adapter~\cite{zhang2024prompt}     & -                 & 256            & 1375.0          & 289.3           \\
        $^*$MiniGPT-4~\cite{zhu2023minigpt}                 & -                 & 256            & 866.5           & 292.1           \\
        $^*$LayerNorm~\cite{zhao2023tuning}       & 325M    & 256  & 929.3 & 254.3 \\
        $^*$LayerNorm-simp.~\cite{zhao2023tuning} & 0.4M    & 256  & 824.3 & 221.1 \\
        $^*$LLaMA-Adapter~\cite{zhang2023llama}             & 14M               & -              & 972.6           & 248.9           \\
        $^*$LaVIN~\cite{luo2024cheap}                       & 5.4M              & 7              & 963.6           & 249.6           \\
        \rowcolor[gray]{0.9}
        $^*$ADEM-VL                                         & 5.5M              & 1              & 966.2           & 270.7           \\
        \bottomrule
    \end{tabular}
    \label{tab:mme}
\end{table*}

\textbf{Instruction following.}
To evaluate our framework on the instruction-following task, we fine-tune a LLaMA-13B model using a combination of the Alpaca-52K and LLaVA-158K datasets, following the implementation of LaVIN~\cite{luo2024cheap}.  
We start by assessing the trained model on the MME benchmark, comparing our results with those reported by LaVIN.
Table~\ref{tab:mme} presents the corresponding results.
Compared to the full fine-tuning method LLaVA, all PEFT methods achieve better performance, with their results showing a trade-off between the two metrics. 
The baseline, Prompt-Aware Adapter, achieves the best overall performance but introduces 256 extra input tokens to the LLM.
Since the LLM accounts for most of the computation in a VL model, this greatly increases both training and inference overhead.  
Our efficiency analysis in Table~\ref{tab:speed} illustrates this point in more detail.  
Compared to the efficient baseline LaVIN, which uses only 7 extra tokens, our proposed framework outperforms across both metrics with only 1 additional token, showcasing its exceptional generalization ability.

To further evaluate the instruction-following model, we assess its performance on additional image understanding tasks, comparing it against both full training methods and PEFT approaches.  
As shown in Table~\ref{tab:more}, across all four image question-answering tasks and benchmark toolkits, our LLaMA-13B-based ADEM-VL model achieves performance comparable to full training methods.  
Although there is a marginal performance gap, our approach involves significantly fewer trainable parameters than full training methods, greatly reducing the resource consumption for fine-tuning VL models.  
When compared with PEFT methods, ADEM-VL with LLaMA-13B delivers the best performance with a similar number of trainable parameters, further highlighting its effectiveness and efficiency.

\begin{table*}
    \renewcommand\tabcolsep{10pt}
    \centering
    \small
    \caption{Comparison among different VL models on more image understanding tasks. $^\ast$Baseline results evaluated through our implementation using the official checkpoint.}
    \begin{tabular}{l|cc|cc|cc}
        \toprule
        \multirow{2}{*}{Method} & \multicolumn{2}{c|}{\#Param} & \multicolumn{2}{c|}{Image QA} & \multicolumn{2}{c}{Benchmark} \\
                                                       & Trainable & LLM  & VQAv2         & GQA           & MMB           & MMMU          \\
        \midrule
        \multicolumn{2}{l}{\emph{Full training methods}}                                                                                  \\
        LLaVA~\cite{liu2024visual}                     & 13B       & 13B  & -             & -             & 34.1          & 32.3          \\
        mPLUG-Owl2~\cite{ye2024mplug}                  & 8.2B      & 8.2B & 79.4          & 56.1          & 64.5          & -             \\
        InternLM-XComposer2~\cite{dong2024internlm}    & 7B        & 7B   & -             & -             & 79.6          & 42.0          \\
        MoE-LLaVA-1.6B$\times$4-Top2~\cite{lin2024moe} & 6.4B      & 6.4B & 76.7          & 60.3          & 60.2          & -             \\
        \midrule
        \multicolumn{2}{l}{\emph{PEFT methods}}                                                                                           \\
        MiniGPT-4~\cite{zhu2023minigpt}                & -         & 13B  & -             & -             & 23.0          & -             \\
        LaVIN~\cite{luo2024cheap}                      & 5.4M      & 13B  & 68.6$^\ast$   & 48.8$^\ast$   & 56.7$^\ast$   & 35.0$^\ast$   \\
        \rowcolor[gray]{0.9}
        ADEM-VL                                        & 4.5M      & 7B   & 71.7          & 52.4          & 52.4          & 34.2          \\
        \rowcolor[gray]{0.9}
        ADEM-VL                                        & 5.5M      & 13B  & \textbf{73.5} & \textbf{56.0} & \textbf{58.4} & \textbf{38.3} \\
        \bottomrule
    \end{tabular}
    \label{tab:more}
\end{table*}

\begin{table*}
    \renewcommand\tabcolsep{4pt}
    \centering
    \small
    \caption{Training and inference speed of different approaches. Memory-saving or speed-up approaches such as checkpointing and flashattention are not adopted. FLOPs are estimated for generating a single new token with a text sequence length of 256. Experiments on COCO captioning and instruction-following were not implemented in the original papers of LLaVA-LoRA and MemVP, so the overall training time for these tasks is unavailable.}
    \begin{tabular}{lccc|cc|ccc}
        \toprule
        \multirow{2}{*}{Method}         & \multicolumn{2}{c}{\#Param} &\multirow{2}{*}{FLOPs}& \multicolumn{2}{c|}{\#Time (s/batch)} & \multicolumn{3}{c}{\#Overall training time (GPU Hours)} \\
                                        & T.   & LLM &         & Training & Inference & ScienceQA & COCO caption & Instruction \\
        \midrule
        LLaVA-LoRA~\cite{jie2024memory} & 4.4M & 7B  & 110.44T & 0.49     & 3.42      & 8.8       & -            & -           \\
        LaVIN~\cite{luo2024cheap}       & 3.8M & 7B  & 56.19T  & 0.39     & 2.06      & 6.8       & 12.7         & 211.4       \\
        MemVP~\cite{jie2024memory}      & 3.9M & 7B  & 54.81T  & 0.28     & 1.88      & 5.1       & -            & -           \\
        MemVP~\cite{jie2024memory}      & 5.5M & 13B & 132.76T & 0.46     & 3.07      & 8.1       & -            & -           \\
        \rowcolor[gray]{0.9}
        ADEM-VL                         & 4.5M & 7B  & 54.93T  & 0.25     & 1.86      & 4.3       & 8.0          & 134.8       \\
        \rowcolor[gray]{0.9}
        ADEM-VL                         & 5.5M & 13B & 133.26T & 0.39     & 2.97      & 6.9       & 12.5         & 212.9       \\
        \bottomrule
    \end{tabular}
    \label{tab:speed}
\end{table*}

\textbf{Efficiency.}
To demonstrate the superiority of our method in delivering more efficient VL models across various PEFT methods, we report the FLOPs required to generate one token in Table~\ref{tab:speed}.
The FLOPs is measured under a text sequence length of 256, which corresponds to a total context length of 512 when 256 additional image tokens are used for methods that fuse VL information at the input stage.  
Compared to the most effective baseline, our method incurs only a slightly higher FLOPs, primarily due to the extended sequence length introduced by the [CLS] token.  
As demonstrated in our ablation study, this token significantly enhances performance while adding negligible computational overhead.

Additionally, we evaluate the training and inference speed of fine-tuned models and compare the results with existing approaches. Following the configuration in~\cite{jie2024memory}, we set the batch size to 4 during training and 64 during inference, using NVIDIA A800 GPUs to measure the speed.
The evaluated results showing that our trained model achieves the lowest training and inference latency when the parameter size in the original LLMs is the same. 
The overall training time further demonstrates the efficiency of our proposed method. For instance, the GPU time required by the ADEM-VL framework to fine-tune a 13B parameter model is even lower than that of LaVIN when fine-tuning a 7B model.

In Table~\ref{tab:main} and Table~\ref{tab:caption}, the results demonstrate that our method outperforms all other PEFT approaches while maintaining higher efficiency. For example, in the image captioning task, our proposed method surpasses LaVIN by 0.7 and 1.5 points in BLEU-4 and CIDEr scores, respectively, while being 36\% faster in training and 10\% faster in inference. This indicates that our design is highly effective for tuning efficient VL models without sacrificing performance. Although the performance is slightly lower than some input-stage fusion approaches, which require significantly more computational resources, our approach offers a promising alternative for achieving intermediate-layer fusion. It is more efficient than input-stage counterparts as it avoids the quadratically increased computational burden associated with longer input sequences. We believe our framework can inspire future research in designing more efficient VL models through intermediate-layer fusion strategies.

\begin{table*}
    \renewcommand\tabcolsep{3pt}
    \centering
    \small
    \caption{Ablation study of each module in our ADEM-VL framework with LLaMA-7B as the language model.}
    \begin{tabular}{lc|ccc|ccc|cc|c}
        \toprule
        \multirow{2}{*}{Setting} & \multirow{2}{*}{\#Trainable} & \multicolumn{3}{c|}{Subject} & \multicolumn{3}{c|}{Context Modality} & \multicolumn{2}{c|}{Grade} & \multirow{2}{*}{Average} \\
                           &      & NAT            & SOC            & LAN            & TXT            & IMG            & NO             & G1-6           & G7-12          &                \\
        \midrule
        Baseline           & 3.4M & 93.49          & 95.05          & 88.21          & 92.85          & 91.28          & 90.92          & 92.50          & 92.35          & 92.45          \\
        + [cls] token      & 4.0M & 93.70          & 95.00          & 88.46          & 93.19          & 91.85          & 90.63          & 92.37          & 93.05          & 92.61          \\
        + Parameter-free xattn & 4.0M & 94.60          & \textbf{95.65} & 89.00          & 94.56          & 93.19          & 90.89          & 93.42          & 93.27          & 93.37          \\
        + Multiscale VP    & 4.5M & 95.10          & 95.50          & 88.50          & 94.87          & 93.48          & 90.66          & 93.61          & 93.21          & 93.47          \\
        + Adaptive fusion  & 4.5M & \textbf{95.52} & 95.39          & \textbf{89.18} & \textbf{95.36} & \textbf{93.95} & \textbf{90.94} & \textbf{93.87} & \textbf{93.80} & \textbf{93.85} \\
        \bottomrule
    \end{tabular}
    \label{tab:ablation:main}
\end{table*}

\begin{table}
    \renewcommand\tabcolsep{12pt}
    \centering
    \small
    \caption{Comparison of different locations for inserting cross-attention modules with LLaMA-7B as the language model. ``Query from'' indicates which features of the language model serve as inputs to the cross-attention modules, while ``Add to'' indicates where the output of these modules is fused into the features of the language model by addition.}
    \begin{tabular}{cc|c}
        \toprule
        Query from  & Add to      & Average        \\
        \midrule
        MHSA (in)~~ & MHSA (in)~~ & 92.19          \\
        MHSA (in)~~ & MHSA (out)  & 93.18          \\
        MHSA (out)  & MHSA (out)  & 92.00          \\
        MLP (in)~~  & MLP (in)~~  & 91.77          \\
        MLP (in)~~  & MLP (out)   & \textbf{93.85} \\
        MLP (out)   & MLP (out)   & 92.27          \\
        \bottomrule
    \end{tabular}
    \label{tab:ablation:xattn_loc}
\end{table}

\begin{table}
    \centering
    \small
    \caption{Comparison of different non-parameterized linear projection in Equation~\ref{eq:xattn_ours} with LLaMA-7B as the language model.}
    \begin{tabular}{cc|c}
        \toprule
        Projection      & formula                       & Average        \\
        \midrule
        None            & $x\to x$                      & 92.16               \\
        Softmax         & $x\to \text{softmax}(x)$      & 79.42          \\
        ReLU            & $x\to \text{relu}(x)$         & 91.99          \\
        ELU             & $x\to \text{elu}(x)$          & 92.45          \\
        SiLU            & $x\to \text{silu}(x)$         & \textbf{93.85} \\
        SiLU (positive) & $x\to \text{silu}(x)-\min(x)$ & 38.58          \\
        \bottomrule
    \end{tabular}
    \label{tab:ablation:xattn_func}
\end{table}

\subsection{Ablation study}\label{sec:exp:ablation}

To examine the effectiveness of each component in our proposed framework, we conduct a series of ablation studies.
We begin with a coarse look at the whole framework and then investigate each module more carefully.

\textbf{Ablation of the overall framework.}
The proposed framework consists of three main modules: parameter-free cross-attention, multiscale visual prompts, and adaptive multimodal fusion. We study the effectiveness of each module by designing ablation experiments on the ScienceQA dataset with LLaMA-7B and report the results in Table~\ref{tab:ablation:main}.
The baseline result, obtained without any of the three modules, achieves an average accuracy of 92.45\%. 
In this setting, we adopt the cross attention in Equation~\ref{eq:xattn_ours} with identity projection $\phi: x \to x$ as the projection function.
On this basis, we first introduce the [cls] token into the input space, which slightly improves performance to 92.61\% while introducing minimal additional resource consumption. Then, we adopt the design in our parameter-free cross-attention by using SiLU activation as the projection, which helps the trained model achieve an average accuracy of 93.37\%.
Further, we introduce multiscale visual information into the multimodal fusion process, which helps the accuracy reach 93.47\%.
Finally, by dropping the least important visual features based on the adaptive fusion module, the model achieves an average accuracy of 93.85\%.
These results demonstrate that each component in our proposed framework contributes to improved performance of fine-tuned VL models. The combination of all components leads to the best result, demonstrating the effectiveness of our design.

\textbf{Locations for fusion.}
Cross-attention modules can be placed at different parts of the pretrained LLM for multimodal fusion.
Previous works mainly position them either before or after the self-attention modules~\cite{wadekar2024evolution}.
Here, we consider a more general setting by decoupling the location of obtaining input and fusing output of cross-attention modules, allowing both the head and tail of the multilayer perceptron (MLP) module as possible positions.
Results in Table~\ref{tab:ablation:xattn_loc} compare different location configurations for fusion.
From the results, simply placing cross-attention before or after self-attention or MLP modules is not as effective as positioning them to span these modules, i.e., input and output being located at the head and the tail of these modules, respectively.
Compared to spanning across the self-attention module, making the cross-attention modules use the same input as the MLP modules and fusing them at the output layer leads to the best performance.

\textbf{Non-parameterized projection.}
In the parameter-free cross-attention module, a fixed, non-parameterized projection is adopted to replace the learnable weight matrices and softmax operation, reducing both parameters and computational complexity.
To investigate the impact of different projection options, we compare several choices in Table~\ref{tab:ablation:xattn_func}.
From the results, the identity projection achieves an average accuracy of 92.16\%. When applying more complex projections, there is no guarantee of better performance, such as softmax and ReLU. In contrast, when adopting SiLU as the projection, the accuracy becomes 93.85\%.
Since the theoretical analysis result requires the output space of projection to be in $\mathbb{R}_+$, we also consider a variant of SiLU that satisfies this by subtracting the minimum value of the input feature. However, this projection leads to significantly worse performance, indicating such a requirement is not necessary in practice.

\textbf{Multiscale visual prompts.}
Multiscale visual features are adopted as prompts in our framework. We compare the impact of different scales and approaches for obtaining such features through experiments and present the results in Table~\ref{tab:ablation:pool}.
We first examine the impact of scale by using two-dimensional average pooling as the downsampling approach.
We apply downsampling with kernel sizes of 2 and 4, resulting in 64 and 16 downsampled visual features, respectively.
When only visual features from a single scale are used, downsampling to either 64 or 16 leads to lower performance compared to using the full set of 256 visual features, with performance degradation becoming more pronounced as the number of visual features decreases.
When multiscale features are obtained by concatenating multiple single-scale features, using a downsampling kernel size of 2 enhances performance of the model. However, as the kernel size increases to 4, the performance tends to decline in most cases. The only exception occurs when both 2$\times$ and 4$\times$ downsampled features are used together, yielding slightly better results than using either scale alone. We speculate that aggressive downsampling may produce overly coarse features, hindering the model to learn from fine-grained details. Therefore, a significant disparity between scales should be avoided when adopting multiscale visual prompts.
Additionally, we compare max pooling with average pooling under the best configuration of scale. The results demonstrate that average pooling is a more preferable approach for downsampling.

\begin{table}
    \renewcommand\tabcolsep{8pt}
    \centering
    \small
    \caption{Comparison of different downsampling methods and scales in generating multimodal visual prompts with LLaMA-7B as the language model.}
    \begin{tabular}{cc|c}
        \toprule
        Down sample  & Size              & Average        \\
        \midrule
        None         & 256               & 93.70          \\
        Avg. pooling & 64                & 92.82          \\
        Avg. pooling & 16                & 91.65          \\
        Avg. pooling & concat(64,16)     & 93.24          \\
        Avg. pooling & concat(256,16)    & 93.65          \\
        Avg. pooling & concat(256,64)    & \textbf{93.85} \\
        Avg. pooling & concat(256,64,16) & 93.59          \\
        Max pooling  & concat(256,64)    & 93.55          \\
        \bottomrule
    \end{tabular}
    \label{tab:ablation:pool}
\end{table}

\begin{figure*}
    \centering
    \includegraphics[width=\textwidth]{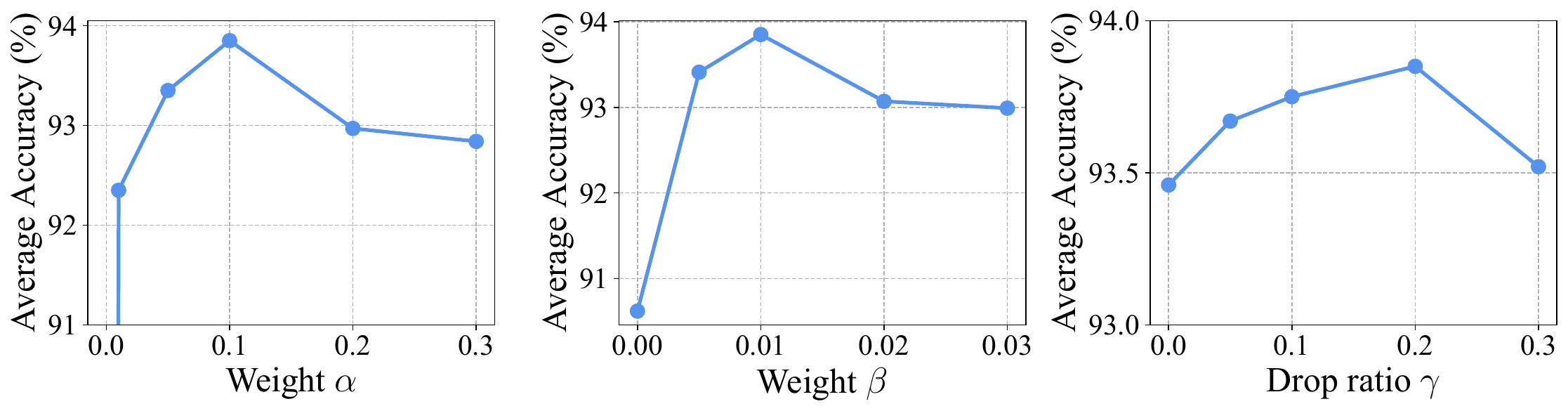}
    \caption{Comparison of different hyperparameter settings in the ADEM-VL with LLaMA-7B as the language model.}
    \label{fig:hyperparameter}
\end{figure*}

\begin{figure*}
    \centering
    \includegraphics[width=\textwidth]{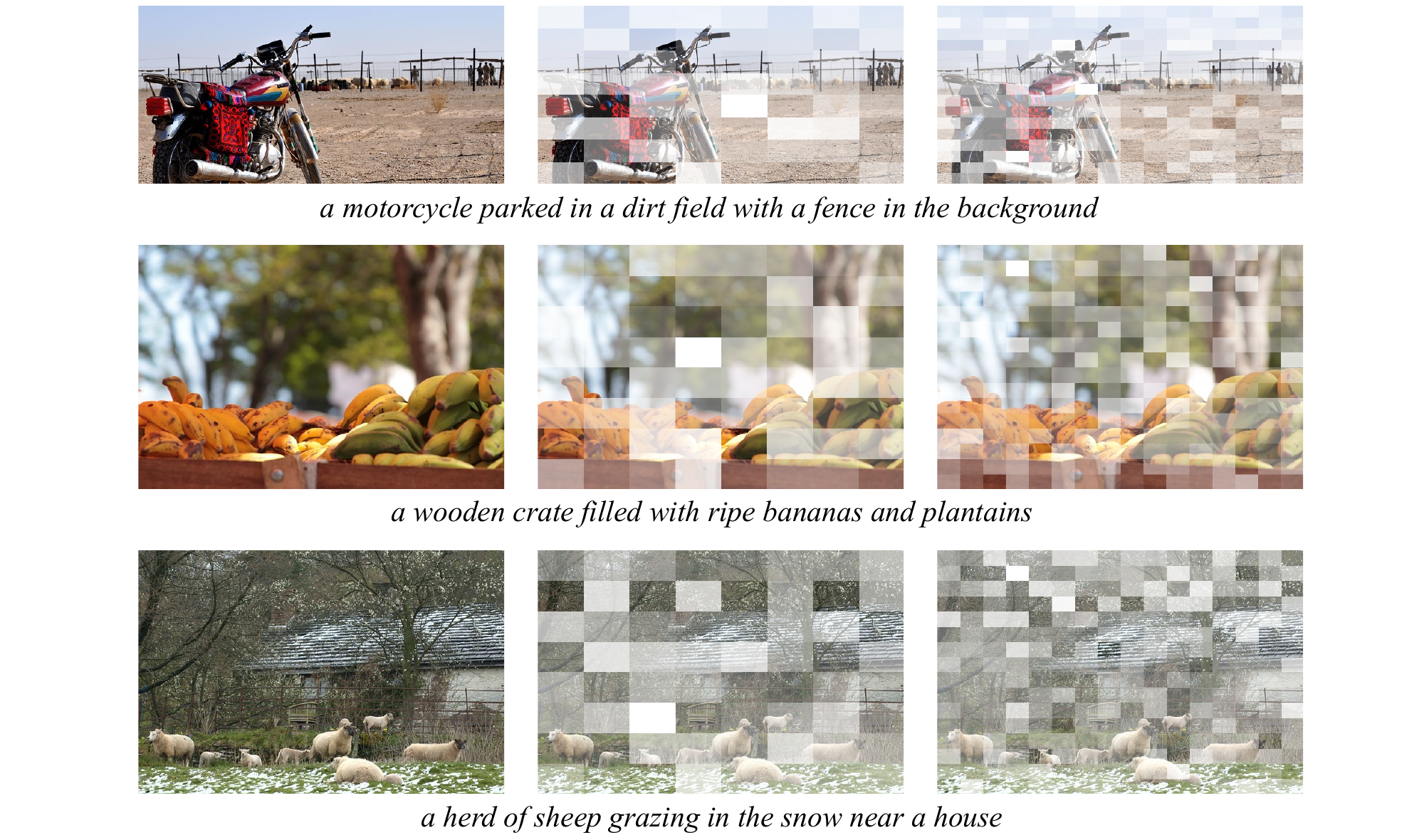}
    \caption{Visualization of image captioning results with LLaMA-7B. In each row, the left figure is the original image, while the middle and right figures demonstrate the dropping decisions for features at two different scales.}
    \label{fig:caption}
\end{figure*}

\textbf{Hyperparameters.}
In our proposed framework, there are three main hyperparameters: weighting parameters $\alpha$ and $\beta$ in Equation~\ref{eq:overall}, and the drop ratio $\gamma$. To investigate the best configurations of these parameters, we compare different settings, and the results are shown in Figure~\ref{fig:hyperparameter}.
The results indicate that the best performance is achieved when $\alpha=0.1$ and $\beta=0.01$.
Regarding the drop ratio $\gamma$, the performance improves as $\gamma$ increases, reaching its peak when $\gamma=0.2$. However, further increasing $\gamma$ shows marginal benefits for adaptive fusion.

\begin{table}
    \renewcommand\tabcolsep{12pt}
    \centering
    \small
    \caption{Integration with different input-stage fusion schemes with LLaMA-7B as the language model.}
    \begin{tabular}{cc|c}
        \toprule
        \multicolumn{2}{c|}{Visual input} & \multirow{2}{*}{Average}        \\
        \#Visual tokens & [cls] token &  \\
        \midrule
        0 & \xmark& 92.97\\
        0& \cmark& \textbf{93.85}\\
        64& \xmark& 92.47\\
        64& \cmark& 92.86\\
        256& \xmark& 89.86\\
        256& \cmark& 90.17\\
        \bottomrule
    \end{tabular}
    \label{tab:ablation:comp}
\end{table}

\textbf{Compatibility with input-stage fusion methods.}
Our method paves the way for building efficient VL models by focusing on feature fusion at intermediate layers.
To examine its compatibility with input-stage fusion methods, we conducted experiments comparing the introduction of visual features at different scales during the input stage.
As shown in Table~\ref{tab:ablation:comp}, the best performance is achieved when only the [cls] token is used as the extra input token alongside text tokens, which is the default setting in our framework. However, when additional visual features are introduced at the input stage, there is a significant performance drop, regardless of the scale of visual features adopted. We speculate that this information has already been incorporated during the intermediate-layer fusion process, and further incorporating it at the input stage may interfere with the fusion, leading to a decline in performance. Additionally, extending the input length increases computational complexity quadratically. Thus, integrating our ADEM-VL framework directly with input-stage fusion approaches may not be an optimal solution.

\subsection{Qualitative results}\label{sec:exp:qual}

\textbf{Image captioning.}
To better understand the adaptive multimodal fusion module in our ADEM-VL framework, we provide quantitative results obtained with our trained model on the image captioning task.
The results are presented in Figure~\ref{fig:caption}. Alongside the original input image, we also visualize the feature-dropping decisions made by the model.
Specifically, for each image feature, each text token has an individual decision on whether to use it at each layer. We collect these decisions and count the frequency of usage for each token. The normalized frequency value is used as the alpha channel for each corresponding patch in the original image, with more transparent patches indicating less importance. Since our framework involves two scales of visual features, we provide the dropping decisions for each scale separately.
As shown in the figure, most of the dropped features are located in the background of the image, such as the tree in the bottom example. However, there are also several patches corresponding to the subject of the image that have been frequently dropped. We speculate that this discrepancy is caused by the different visual perception mechanisms between VL models and humans and will investigate this in depth in future work.

\begin{figure*}
    \centering
    \includegraphics[width=\textwidth]{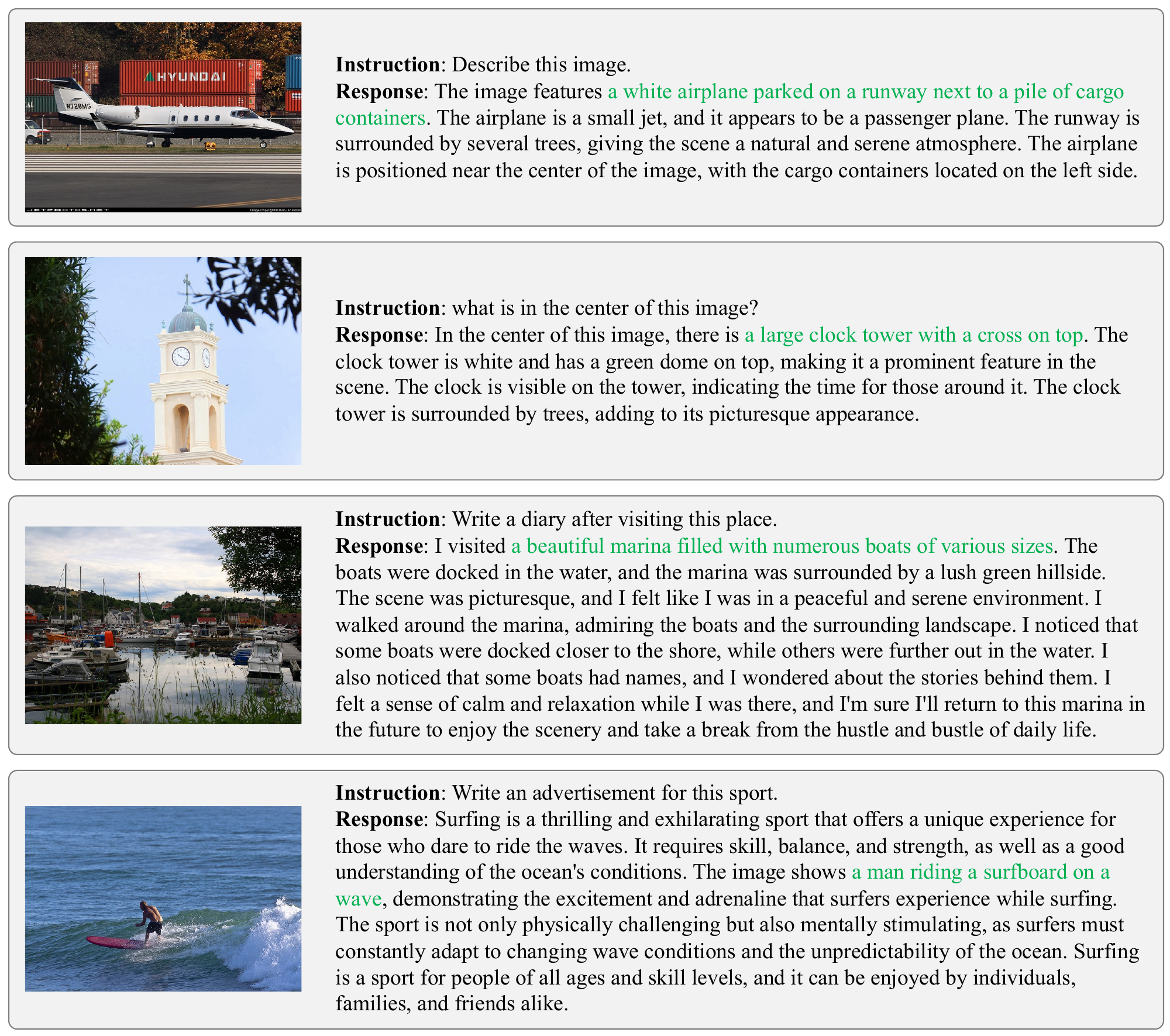}
    \caption{Examples of zero-shot instruction-following tasks with LLaMA-7B.}
    \label{fig:instruction}
\end{figure*}

\textbf{Instruction following.}
We further evaluate the instruction-following data trained model through multimodal chat, where the model responds to an image and a corresponding instruction. Figure~\ref{fig:instruction} presents the results. As shown, the model can recognize the subject in each image and respond correctly to the instruction. Additionally, the model can make proper inferences about the emotions depicted in the image, such as in the last two examples, even when these emotions are not directly demonstrated.
These results showcase that our framework is not only efficient in both training and inference but also highly competitive in practical scenarios.

\section{Conclusion}

Building VL models upon pretrained LLMs has been proven effective but requires considerable computational resources. Facing the absence of efficient VL tuning methods, we propose a framework named ADEM-VL in this paper. 
The proposed framework consists of three main modules: parameter-free cross-attention for efficient multimodal fusion by removing most trainable parameters, multiscale visual prompts for ample visual information by downsampling, and adaptive multimodal fusion for target-focused learning by dynamically dropping useless features.
We conduct experiments on visual question answering, image captioning, and instruction-following tasks, where quantitative results demonstrate the superiority of our proposed method over existing approaches. Additionally, we provide visualization results obtained by our trained model, further demonstrating its effectiveness.
One possible limitation of the proposed framework lies in the adaptive fusion module. The visualization of dropped features does not always align with the order of importance as perceived by humans. Further study should be conducted to improve this module or investigate the core differences between VL models and human perception to further enhance the performance of these models.

\section*{Acknowledgements}

This paper is supported by National Key Research and Development Program of China under No. 2021YFC3300200, and Joint Funds of the National Natural Science Foundation of China No. U2336211.
This paper is supported by National Key Research and Development Program of China under No. 2021YFC3300200, and Joint Funds of the National Natural Science Foundation of China No. U2336211.

\bibliographystyle{IEEEtran}
\bibliography{ref}

\end{document}